\documentclass[lettersize,journal]{IEEEtran}
\pdfoutput=1
\usepackage{amsmath,amsfonts}
\usepackage{algorithm}
\usepackage{array}
\usepackage{textcomp}
\usepackage{stfloats}
\usepackage{url}
\usepackage{verbatim}
\usepackage{graphicx}
\usepackage{cite}
\usepackage{booktabs}
\usepackage{multirow}
\usepackage{makecell}
\usepackage{stackengine}
\usepackage{threeparttable}
\usepackage{xcolor} 
\usepackage{tikz}
\usepackage{soul}
\usepackage[normalem]{ulem}
\usepackage{diagbox}
\usepackage{subfig}
\usepackage{esvect}
\usepackage{algpseudocode}  
\usepackage[colorlinks,
bookmarksopen,
bookmarksnumbered,
citecolor=blue,
linkcolor=red,
urlcolor=red]{hyperref}
\usepackage{bookmark}
\usepackage{arydshln}
\hypersetup{pdfborder={0 0 0}}

\hypersetup{hidelinks,
	colorlinks=true,
	allcolors=black,
	pdfstartview=Fit,
	breaklinks=true}
	
\definecolor{lime}{HTML}{A6CE39}

\DeclareRobustCommand{\orcidicon}{	
\begin{tikzpicture}
\draw[lime, fill=lime] (0,0)
circle[radius=0.16]
node[white]{{\fontfamily{qag}\selectfont \tiny \.{I}D}};
\end{tikzpicture}
\hspace{-2mm}
}

\foreach \x in {A, ..., Z}{%
\expandafter\xdef\csname orcid\x\endcsname{\noexpand\href{https://orcid.org/\csname orcidauthor\x\endcsname}{\noexpand\orcidicon}}
}

\begin{document}
	
	\title{ViBA: Implicit Bundle Adjustment with Geometric and Temporal Consistency for Robust Visual Matching}
	
	\author{
		Xiaoji Niu\hspace{-1.5mm}\orcidF{}, ~\IEEEmembership{Member,~IEEE}, Yuqing Wang\hspace{-1.5mm}\orcidA{}, ~\IEEEmembership{Graduate Student Member, IEEE}, Yan Wang*\hspace{-1.5mm}\orcidB{}, Hailiang Tang\hspace{-1.5mm}\orcidC{}, and Tisheng Zhang\hspace{-1.5mm}\orcidG{},~\IEEEmembership{Member,~IEEE}
		\thanks{Yuqing Wang is with the GNSS Research Center, Wuhan University,	Wuhan 430079, China, and also with the Electronic Information School, Wuhan University, Wuhan 430079, China (e-mail: m18071357432@163.com).}
		\thanks{Yan Wang (Corresponding author), Hailiang Tang, are with the GNSS Research Center, Wuhan University, Wuhan 430079, China, also with the Hubei Technology Innovation Center for Spatiotemporal Information and Positioning Navigation (e-mail: thl@whu.edu.cn; wystephen@whu.edu.cn).}
		\thanks{Tisheng Zhang and Xiaoji Niu are with the GNSS Research Center, Wuhan University, Wuhan 430079, China, also with the Electronic Information School, Wuhan University, Wuhan 430079, China, also with Hubei Technology Innovation Center for Spatiotemporal Information and Positioning Navigation, Wuhan 430079, China, and also with Hubei Luojia Laboratory, Wuhan 430079, China (e-mail: zts@whu.edu.cn; xjniu@whu.edu.cn).}
	}

\markboth{Journal of \LaTeX\ Class Files,~Vol.~14, No.~8, August~2021}%
{Shell \MakeLowercase{\textit{et al.}}: A Sample Article Using IEEEtran.cls for IEEE Journals}

\IEEEpubid{}
\maketitle

\begin{abstract}
Most existing image keypoint detection and description methods rely on datasets with accurate pose and depth annotations, limiting scalability and generalization and often degrading navigation and localization performance. We propose ViBA, a sustainable learning framework that integrates geometric optimization with feature learning for continuous online training on unconstrained video streams. Embedded in a standard visual odometry pipeline, it consists of an implicitly differentiable geometric residual framework: (i) an initial tracking network for inter-frame correspondences, (ii) depth-based outlier filtering, and (iii) differentiable global bundle adjustment that jointly refines camera poses and feature positions by minimizing reprojection errors. By combining geometric consistency from BA with long-term temporal consistency across frames, ViBA enforces stable and accurate feature representations. We evaluate ViBA on EuRoC and UMA datasets. Compared with state-of-the-art methods such as SuperPoint+SuperGlue, ALIKED, and LightGlue, ViBA reduces mean absolute translation error (ATE) by 12–18\% and absolute rotation error (ARE) by 5–10\% across sequences, while maintaining real-time inference speeds (FPS 36–91). When evaluated on unseen sequences, it retains over 90\% localization accuracy, demonstrating robust generalization. These results show that ViBA supports continuous online learning with geometric and temporal consistency, consistently improving navigation and localization in real-world scenarios.
\end{abstract}


\section{Introduction}\label{I}
\IEEEPARstart{V}{isual} odometry (VO) \cite{vo} and visual–inertial odometry (VIO) \cite{vins} estimate camera motion by minimizing geometric residuals over tracked visual features across image sequences. As a result, the accuracy and robustness of VO/VIO systems critically depend on the ability to establish long-term, reliable, and geometrically consistent feature correspondences \cite{orb, orb2,ogi-slam2,ptam}. Failures in feature tracking or matching directly lead to degraded pose estimation, scale drift, or even system breakdown, particularly in challenging real-world environments, such as involving rapid motion, illumination changes, or weak textures \cite{dso,6907054,9070681}.

Classical feature-based methods, such as handcrafted keypoint detectors and descriptors \cite{Scale-Invariant, RPI-SURF, shi-tomasi}, have demonstrated reliable performance and high efficiency in many VO and SLAM systems. However, their performance is inherently limited by fixed design choices and heuristic assumptions, which restrict adaptability to diverse sensing conditions. Recent learning-based approaches have significantly improved feature repeatability and matching accuracy by leveraging large-scale datasets and deep neural networks \cite{SuperPoint, aliked, ASLFeat}. Despite their success, most existing keypoint detection and description methods are trained independently of the downstream geometric optimization process \cite{dedode}. In particular, supervision is commonly provided through ground-truth correspondences, pose annotations, or proxy objectives that are only loosely related to the reprojection errors minimized in VO/VIO pipelines. Although several recent works attempt to improve long-term descriptor consistency\cite{gim} or replace photometric invariance in optical flow formulations \cite{let-net}, learned features are still not explicitly optimized for long-term geometric consistency. As a result, their effectiveness in navigation and localization tasks remains limited.

Several recent works have explored incorporating geometric constraints into feature learning, including epipolar consistency \cite{superglue,Correspondences}, photometric losses\cite{8100182,zhan2018unsupervisedlearningmonoculardepth}, and local pose supervision\cite{8100132,Ego-Motion}. While these approaches partially bridge the gap between learning and geometry, they typically rely on simplified or local formulations and do not account for the global optimization procedures that define modern VO and SLAM systems. In practice, state-of-the-art pipelines rely on iterative global optimization techniques, such as bundle adjustment (BA)\cite{ba}, to jointly refine camera poses and feature estimates. However, the implicit and iterative nature of bundle adjustment makes it difficult to integrate directly into end-to-end feature learning frameworks \cite{ba-net}, preventing learned features from being optimized with respect to the true system-level objectives of VO/VIO pipelines.

Implicit differentiation provides a principled mechanism for backpropagating gradients through optimization problems defined by equilibrium conditions, without requiring unrolling of iterative solvers \cite{dnls}. This property makes it particularly attractive for integrating global geometric optimization into learning-based perception systems. Although implicit differentiation has been successfully applied to selected problems in robotics, such as autonomous flight of unmanned aerial vehicles \cite{agileflight} and robot control \cite{dojo}, its potential for feature matching and long-term tracking in VO/VIO systems remains largely underexplored. In particular, how to leverage implicit differentiation to couple feature learning with long-term geometric consistency, while maintaining robustness and efficiency in real-world video streams, remains an open challenge.

In this paper, we propose a geometry-aware feature learning framework that leverages implicit differentiation to optimize feature representations for VO/VIO objectives. The method is tightly integrated into a standard visual odometry pipeline and supports continuous online learning from unconstrained video streams. At its core, the framework incorporates an implicitly differentiable geometric residual formulation that enables reprojection errors from a global bundle adjustment (BA) module to be backpropagated to both the feature tracking and feature extraction networks, refining feature representations directly through geometric optimization.

Through repeated propagation of reprojection residuals via implicit optimization layers, the framework enforces self-consistency and long-term geometric consistency in the learned features. Unlike prior approaches that rely on static datasets with pose and depth annotations, our method enables continuous online learning directly from arbitrary video streams.

\begin{figure*}
	\centering
	{\includegraphics[width=0.9\linewidth]{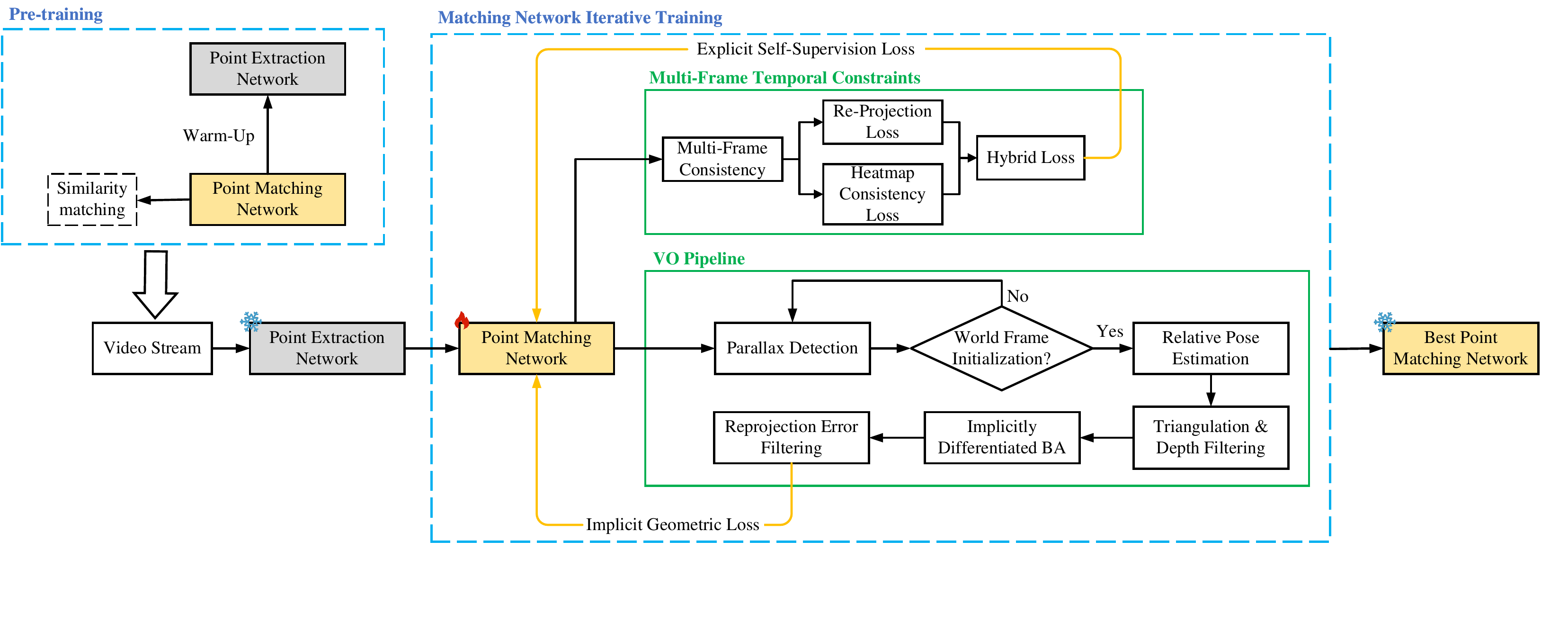}}
	
	\caption{\textbf{The pipeline of the proposed method}. The snowflake denotes a frozen network, the flame indicates an activated network, and the spark associated with the loss marks the transition from frozen to activated.}
	\label{Fig:1}
\end{figure*}

The structure of the proposed method is illustrated in Fig.~\ref{Fig:1}. The framework consists of a Feature Tracking module and a Feature Extraction module. The Feature Tracking module comprises three sequential components: initial feature tracking, geometry perception, and implicit geometric optimization. Given an input video stream, the initial tracking component establishes sparse inter-frame correspondences using a learned tracking network, producing preliminary feature trajectories and associated local descriptors. These correspondences are then passed to the geometry perception component, which performs depth-based filtering together with conventional triangulation and pose estimation to recover camera motion and sparse scene structure. Based on the estimated poses and 3D points, a Global Bundle Adjustment (BA) module is subsequently activated. The BA module is formulated as an implicitly differentiable optimization layer that jointly refines camera poses and scene structure by minimizing the global reprojection error, and propagates the resulting geometric residuals back to the feature tracking process.

To further improve feature robustness, the framework enforces temporal consistency across multiple frames by maintaining coherence of feature trajectories. This combination of geometric and temporal consistency ensures that the learned features are stable and self-consistent over time, supporting reliable visual odometry and localization without relying on alternating training or online network updates.

Overall, the main contributions of this paper are as follows:
\begin{itemize}

	\item We propose an online geometry-aware feature learning framework that integrates an implicitly differentiable global bundle adjustment into a standard VO/VIO pipeline, enabling feature matching to be directly optimized with respect to geometric reprojection objectives.
	
	\item We enforce local temporal consistency by refining feature descriptors based on their stability and correspondence across adjacent frames, ensuring features are consistent over time without relying on global alignment.
	
	\item The proposed framework supports self-supervised online training directly from unconstrained video streams, without requiring pre-collected datasets or explicit pose and depth annotations, and improves robustness and accuracy in navigation and localization tasks.
\end{itemize}

\section{Related Work}\label{II}
\subsection{Feature-Based VO/VIO}\label{II-A}
Feature-based methods are a cornerstone of visual odometry (VO) and visual–inertial odometry (VIO) systems. Classical approaches employ handcrafted keypoint detectors and descriptors, such as SIFT \cite{sift}, SURF \cite{surf}, and ORB \cite{orb-feature}, to extract sparse and repeatable features for inter-frame matching and tracking. When combined with geometric estimation and optimization techniques, including pose estimation, triangulation, and bundle adjustment (BA), these methods have been widely adopted in VO and SLAM systems. However, their performance is inherently constrained by fixed design choices and heuristic assumptions, limiting robustness and adaptability in complex real-world environments.

To overcome these limitations, recent studies have explored learning-based keypoint detection and description methods. Deep features such as SuperPoint \cite{SuperPoint}, D2-Net \cite{D2NET}, and R2D2 \cite{r2d2} have demonstrated improved robustness and matching accuracy under challenging imaging conditions, and have been incorporated as front-end components in modern SLAM and VO systems. For example, Qu et al. \cite{dk-slam} proposed DK-SLAM, which integrates a SuperPoint-based front-end into a monocular SLAM framework, enabling learned keypoint detection, tracking, and loop closure for improved robustness. Li et al. \cite{dx-slam} incorporated learning-based front-ends such as SuperPoint and D2-Net into a classical SLAM system, leveraging learned local and global CNN features to enhance tracking and loop closure while maintaining real-time performance on CPU-only platforms. Despite these advances, most learning-based features are trained using surrogate objectives, such as ground-truth correspondences or homography-based supervision, rather than being explicitly optimized for downstream geometric objectives (e.g., reprojection error minimization), which ultimately determine VO and VIO performance. This mismatch limits their effectiveness in long-term tracking and geometric consistency.

\subsection{Geometry-Aware Learning for VO/VIO}\label{II-B}
Recent research has aimed to bridge the gap between learned representations and geometric optimization by explicitly integrating downstream objectives into the training process. Instead of relying solely on surrogate supervision, such as ground-truth correspondences, homography constraints, or photometric losses, these methods embed geometric estimation modules—e.g., pose estimation, depth triangulation, or bundle adjustment—directly into the learning pipeline. This design allows the network to learn task-specific and goal-oriented feature or depth representations that are explicitly aligned with the objectives of VO and VIO systems.

For instance, Tang et al. \cite{ba-net} introduced a differentiable bundle adjustment layer to jointly optimize depth and camera poses for structure-from-motion tasks, demonstrating improved reconstruction accuracy. Jan et al. \cite{DeepFactors} proposed a network that encodes full-image depth priors for monocular SLAM, combining learned priors with traditional optimization to maintain long-term geometric consistency. More recently, end-to-end visual odometry systems such as DPVO (Zachary et al. \cite{dpvo}) integrate differentiable pose estimation and iterative optimization, allowing gradients to flow through both motion estimation and feature updates. Similarly, DROID-SLAM (Jia et al. \cite{DROID-SLAM}) unrolls an iterative optimization process to jointly refine camera poses and dense correspondences in a fully differentiable framework, improving trajectory accuracy and loop closure reliability.

Although these geometry-aware learning approaches improve accuracy and robustness by aligning features or depth maps with reprojection and pose objectives, most are trained offline on fixed datasets and focus on dense scene representations rather than directly optimizing sparse feature tracking. As a result, their adaptability to long-term environmental changes, real-time applicability, and robustness in dynamic or low-texture scenarios remain limited.

\subsection{Self-Supervised and Online Adaptation for VO/VIO}\label{II-C}

To further alleviate the reliance on offline supervision, recent research has explored self-supervised and online adaptation strategies for visual odometry and image matching. Unlike approaches that depend on ground-truth depth, pose, or correspondence annotations, self-supervised methods exploit the inherent geometric consistency in multi-view image sequences as supervisory signals. By leveraging label propagation strategies, minimizing photometric reprojection error, and enforcing epipolar constraints, these methods learn feature representations directly from raw image streams without external annotations.

Zhou et al.~\cite{sfmlearner} proposed SfMLearner, which jointly learns monocular depth and relative camera motion from raw video by minimizing photometric view-synthesis error, demonstrating that geometric consistency can serve as effective self-supervision without ground-truth depth or pose annotations. Godard et al.~\cite{monodepth2} revisited self-supervised monocular depth learning and introduced minimum reprojection loss and auto-masking strategies to improve training stability and robustness, showing that carefully designed photometric objectives can significantly enhance depth estimation without ground-truth supervision.

In the context of correspondence learning, Truong et al.~\cite{glunet} introduced GLU-Net, a global-local architecture that combines coarse global matching with local refinement to improve dense correspondence estimation under large viewpoint variations. Wang et al.~\cite{pdcnet} proposed PDC-Net, which incorporates probabilistic dense matching and uncertainty estimation to enhance robustness in challenging scenarios. Researchers have further investigated online adaptation mechanisms to improve robustness under distribution shifts. For example, Shen et al. presented GIM \cite{gim}, a generalizable image matching framework that propagates reliable correspondences across temporally distant frames in Internet videos, significantly improving matching robustness under large viewpoint changes through large-scale label propagation.

Despite these advances, most existing self-supervised or online geometry-aware methods primarily focus on dense depth estimation or global latent representations, rather than directly optimizing sparse feature detection and tracking quality within a geometric optimization pipeline. Moreover, online updates are typically applied to high-level representations outside the core bundle adjustment loop, which may compromise stability and real-time performance. Consequently, a unified framework that enables fully self-supervised, geometry-consistent, and online optimization of sparse feature representations tightly coupled with the BA process remains largely unexplored.

In this work, we address this gap by proposing a self-supervised framework that directly couples sparse feature learning with bundle adjustment objectives, without requiring ground-truth depth or pose annotations. Our method performs online adaptation within the geometric optimization loop, improving long-term tracking stability and robustness while maintaining real-time performance.

\section{PRE-TRAINING}\label{III}
As shown in Fig.~\ref{Fig:1}, during the pre-training stage we focus on two core components: a point extraction network and a point matching network. The extraction network takes the original image as input and produces a dense response map at the original resolution using a fully convolutional architecture, from which high-confidence keypoints are selected as tracking candidates. Given these candidates, the matching network performs data association by comparing local descriptor representations and selecting the most similar points in the target frame.

\begin{figure*}
	\centering
	{\includegraphics[width=0.9\linewidth]{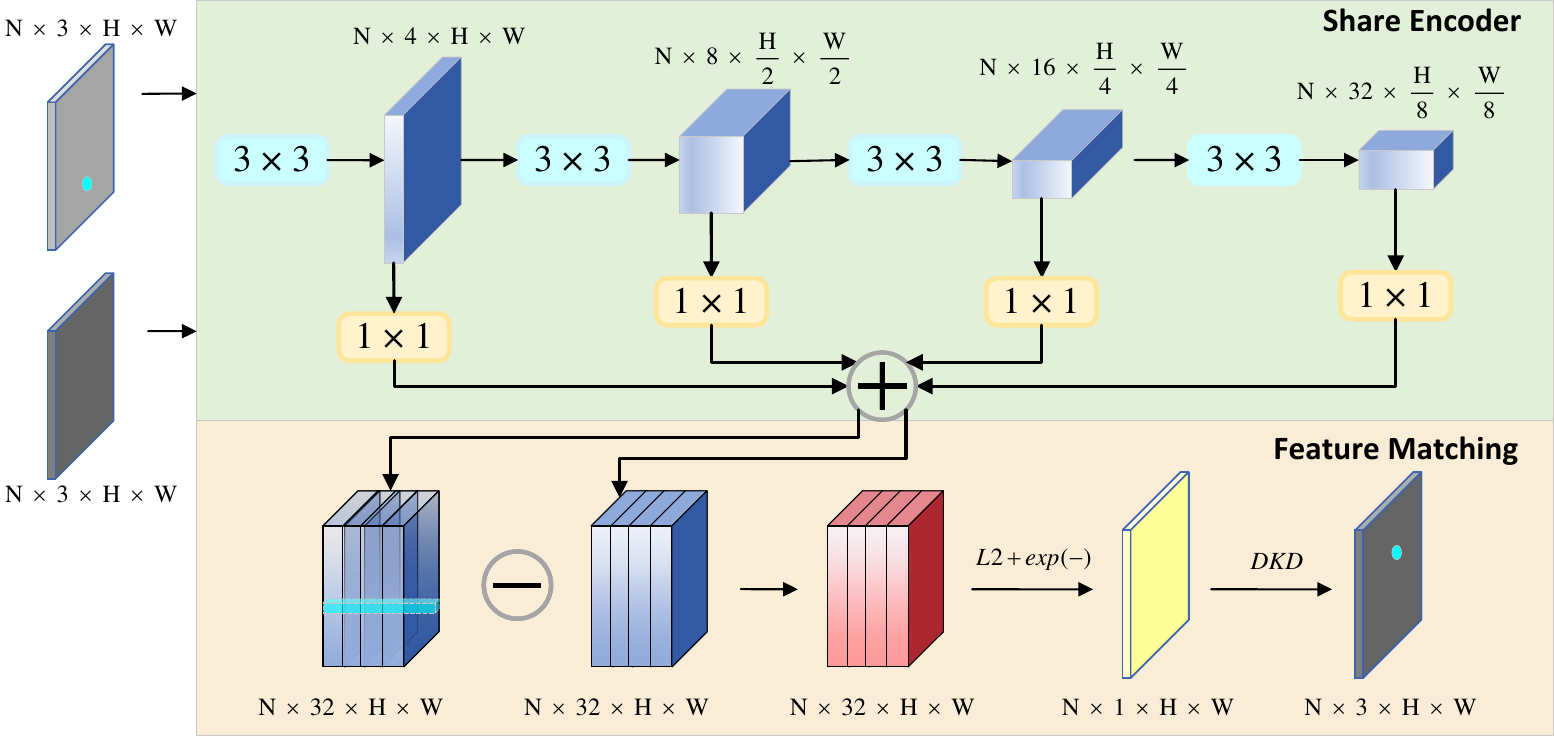}}
	
	\caption{Architecture of the point matching network. It comprises a shared encoder and a feature matching module. Blocks labeled \(1 \times 1\) and \(3 \times 3\) represent residual units with the corresponding convolution kernel sizes. $N$ denotes the number of local image patches.}
	\label{Fig:2}
\end{figure*}

\subsection{Point Matching Network Pre-Training}\label{III-A}
Given an image sequence $\{\mathbf{I}_A, \mathbf{I}_B, \ldots, \mathbf{I}_M\}$, our objective is to select an initial set of keypoints $\mathbf{p}_A$ in frame $\mathbf{I}_A$, and recover their motion trajectories across subsequent frames $\mathbf{I}_N$, where $N = A, \ldots, M$.

To better align the design with visual odometry as the downstream task, we assume short-term temporal continuity of feature motion, i.e., feature points vary smoothly between adjacent frames without abrupt global appearance changes. Accordingly, local image patches centered at detected keypoints are used as inputs to the point matching network, and feature similarity is adopted as the matching criterion (see Fig.~\ref{Fig:2}).

Specifically, given local patches $\mathbf{H}_a$ and $\mathbf{H}_n$ extracted from frames $\mathbf{I}_A$ and $\mathbf{I}_N$ around points $\mathbf{p}_a \in \mathbf{p}_A$ and $\mathbf{p}_n \in \mathbf{p}_N$, a convolutional network with upsampling layers produces low-dimensional dense descriptor maps $\mathbf{D}_a, \mathbf{D}_n \in \mathbb{R}^{H \times W \times 32}$. To obtain the precise descriptor at location $\mathbf{p}_a$, bilinear interpolation is applied to $\mathbf{D}_a$ to extract the corresponding descriptor $\mathbf{d}_{\mathbf{p}_a} \in \mathbb{R}^{32}$. We then compute a similarity difference map $\mathbf{C}$ by comparing $\mathbf{d}_{\mathbf{p}_a}$ with $\mathbf{D}_n$ via element-wise subtraction followed by normalization. Similarly, $\mathbf{d}_{\mathbf{p}_n}$ is compared with $\mathbf{D}_a$ to enforce bidirectional consistency. 

\[
\mathbf{C}(x,y)
=
\exp
\left(
-
\|\mathbf{D}(x,y) - \mathbf{d}_{\mathbf{p}}\|
\right).
\]

Finally, Differentiable Keypoint Detection (DKD) \cite{alike} is applied to detect reliable response peaks, from which correspondence pairs are established.

\subsection{Point Extraction Network Warm-up}\label{III-B}

To stabilize training and provide a reasonable initialization for the point extraction network, we adopt a warm-up strategy based on the pre-trained point matching network. Specifically, we follow a D2-Net-style formulation, where detection and description are jointly derived from shared feature maps. Given the dense descriptor maps produced by the pre-trained matching network, the point extraction network is initialized by activating high-response locations in a D2-Net-like \cite{D2NET} manner. This allows the detector to inherit discriminative properties from the matching network and to focus on repeatable and matchable regions.

It is worth noting that this warm-up procedure does not introduce additional supervision and is not a core contribution of this work. Instead, it serves as a practical initialization strategy to facilitate stable convergence in subsequent self-supervised optimization.

\section{Implicit Differentiation for End-to-End Geometric Optimization}\label{IV}

In this section, we introduce a differentiable geometric optimization framework that enables end-to-end learning of both feature representations and geometric state estimation. We first revisit the classical bundle adjustment formulation and discuss its limitations in modern robotic perception systems. Then, we present a learning-based formulation that integrates neural representation learning with geometric optimization objectives. Finally, we introduce an implicit differentiation strategy that enables gradient propagation through the geometric solver, allowing joint optimization of perception and geometry in a unified framework.

\subsection{Bundle Adjustment}\label{BA}

Given images $\mathbf{I}=\{I_i | i=1\cdots N_i\}$, the geometric state estimation problem in robotic visual perception is typically formulated as a nonlinear optimization problem that jointly estimates camera motion and scene structure. Specifically, let the camera poses be denoted as $\mathbf{T}=\{T_i | i=1\cdots N_i\}$ and the 3D scene landmarks be denoted as $\mathbf{P}=\{p_j | j=1\cdots N_j\}$. The classical geometric bundle adjustment (BA) framework estimates all geometric variables by minimizing the reprojection error between observed image features and projected 3D points.

Formally, the joint optimization problem can be written as

\begin{equation}
	\mathbf{X}^* =
	\arg\min_{\mathbf{X}}
	\sum_{i=1}^{N_i}
	\sum_{j=1}^{N_j}
	\left\|
	\mathbf{e}_{ij}(\mathbf{X})
	\right\| ,
	\label{eq:ba_objective}
\end{equation}
where the geometric reprojection residual is defined as

\begin{equation}
	\mathbf{e}_{ij}(\mathbf{X})
	=
	\pi(T_i, p_j) - q_{ij}.
\end{equation}

Here $\pi(\cdot)$ denotes the camera projection function that maps 3D scene points to image space, and $q_{ij}=[x_{ij}, y_{ij}, 1]^\top$ represents the normalized homogeneous image coordinate of the observed feature. The variable

\begin{equation}
	\mathbf{X} =
	[T_1, T_2, \cdots, T_{N_i},
	p_1, p_2, \cdots, p_{N_j}]^\top
	\label{eq:3}
\end{equation}
contains all geometric parameters to be optimized.

In practice, nonlinear bundle adjustment problems are commonly solved using second-order optimization algorithms such as the Levenberg-Marquardt (LM) algorithm. At each iteration, LM computes an optimal update $\Delta \mathbf{X}^*$ by solving
\begin{equation}
	\Delta \mathbf{X}^* =
	\arg\min
	\left\|
	J(\mathbf{X})\Delta \mathbf{X}
	+
	E(\mathbf{X})
	\right\|^2
	+
	\lambda
	\left\|
	D(\mathbf{X})\Delta \mathbf{X}
	\right\|^2,
\end{equation}
where
\begin{equation}
	E(\mathbf{X}) =
	[\mathbf{e}_{11}(\mathbf{X}),
	\mathbf{e}_{12}(\mathbf{X}),
	\cdots,
	\mathbf{e}_{N_iN_j}(\mathbf{X})]
\end{equation}
is the stacked residual vector, $J(\mathbf{X})$ denotes the Jacobian matrix of residuals with respect to the state variables, and $D(\mathbf{X})$ is a diagonal damping matrix typically constructed from the square root of the approximate Hessian $J^\top J$. The non-negative parameter $\lambda$ controls the regularization strength and balances convergence stability and optimization speed.

The special block-sparse structure of the normal equation system $J^\top J$ enables efficient optimization using techniques such as Schur complement decomposition, which significantly reduces computational complexity when optimizing large-scale structure-from-motion problems.

Although classical geometric BA has served as the gold standard for structure-from-motion and visual odometry for decades, it relies heavily on reliable feature extraction and matching. In particular, only sparse image information corresponding to predefined feature types, such as corners, blobs, or line segments, is utilized. This design introduces two fundamental limitations.

First, feature matching between frames is required to establish geometric correspondences. However, feature matching inevitably introduces outlier correspondences due to repetitive textures, illumination changes, or dynamic scene content. Robust estimation techniques such as RANSAC are typically employed for outlier rejection, yet they cannot guarantee globally optimal matching results in highly ambiguous environments.

Second, sparse feature-based methods do not fully exploit dense image information. Large portions of the image that contain geometric cues but do not correspond to detected features are ignored, limiting the robustness of state estimation in low-texture environments.

These limitations have motivated the development of direct visual odometry methods, which attempt to eliminate explicit feature matching by directly minimizing photometric consistency across images. In direct methods, the photometric error is defined as
\begin{equation}
	e_{ij}^p(\mathbf{X})
	=
	I_i(\pi(T_i, d_j q_j))
	-
	I_1(q_j),
\end{equation}
where $d_j$ denotes the depth value associated with pixel $q_j$, and $d_j q_j$ reconstructs the corresponding 3D point in camera coordinates.

In this formulation, the optimization variables become
\begin{equation}
	\mathbf{X} =
	[T_1, T_2, \cdots, T_{N_i},
	d_1, d_2, \cdots, d_{N_j}]^\top.
\end{equation}

Direct methods have the advantage of exploiting dense pixel information with sufficient image gradients, enabling superior performance in low-texture scenes. However, these methods also introduce additional challenges. They are highly sensitive to initialization quality because photometric error functions are highly non-convex. In addition, photometric consistency assumptions are easily violated under varying exposure settings, automatic white balance adjustments, or dynamic lighting conditions. Furthermore, direct methods are more vulnerable to dynamic scene interference caused by moving objects.

These limitations motivate the development of learning-based differentiable geometric optimization frameworks that can jointly model feature representation and geometric constraints in a unified optimization paradigm.

\subsection{Implicit Differentiation of the Geometric Solver}

To achieve this goal, we revisit the nonlinear least-squares formulation of geometric state estimation from a probabilistic perspective. In visual state estimation, feature observations are assumed to be conditionally independent given the geometric state, and corrupted by Gaussian noise. Under this assumption, maximum likelihood estimation reduces to a nonlinear least-squares problem that enforces consistency between predicted visual observations and geometric constraints.

Instead of introducing additional hand-crafted optimization components, our objective is to improve the geometric reliability of existing tracking networks while preserving their representational capacity. Specifically, the tracking model is treated as a learned observation function that produces feature correspondences from input images, while the geometric solver enforces spatial and temporal consistency through structured geometric optimization.

However, enforcing geometric consistency directly within the tracking pipeline is non-trivial. In conventional pipelines, the tracking network and geometric estimation module are optimized independently using distinct objective functions. The tracker is usually trained with appearance-based supervision, while the geometric solver minimizes reprojection errors without influencing feature representation learning. This separation prevents the learned features from adapting to downstream geometric constraints.

To address this limitation, we introduce an implicitly differentiable bundle adjustment (BA) formulation that refines tracking predictions under geometric supervision. Within this framework, the tracking network is not replaced by the geometric solver; instead, it is optimized through geometric consistency signals, enabling the model to produce correspondences that are both visually discriminative and geometrically consistent.

\begin{figure}
	\centering
	{\includegraphics[width=0.9\linewidth]{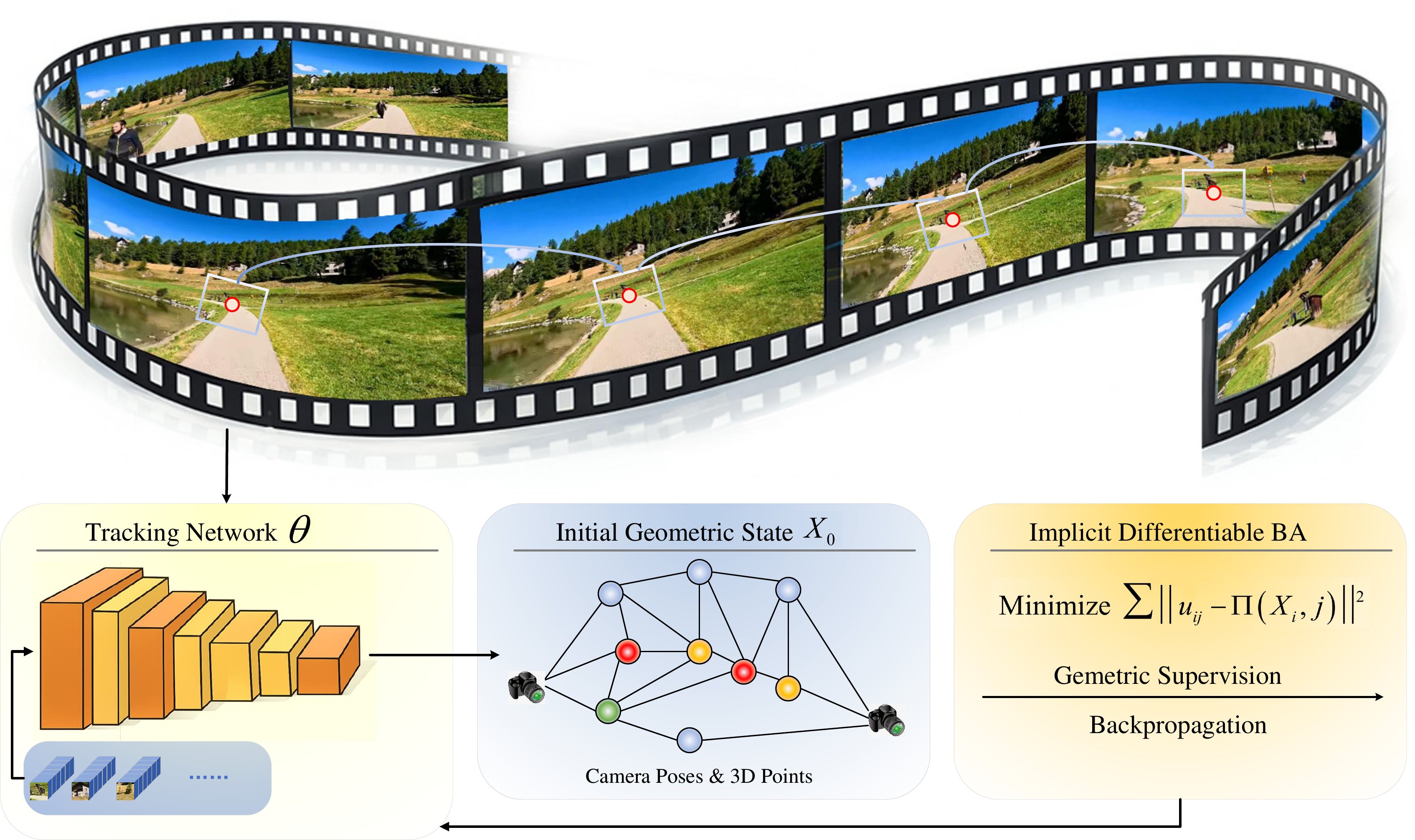}}
	
	\caption{Feature tracking and geometric initialization. The tracking network $\theta$ predicts feature observations from input images to form multi-frame feature tracks, which are then used to initialize the geometric state (camera poses and feature depths) for subsequent optimization.}
	\label{Fig:3}
\end{figure}

Let the tracking network be parameterized by $\theta$, which predicts feature observations from the input image, as illustrated in Fig.~\ref{Fig:3}. The learned observation model is expressed as

\begin{equation}
	\mathbf{u}_{ij} = f_\theta(\mathbf{I}_i, j),
\end{equation}
where $\mathbf{u}_{ij}$ denotes the predicted image coordinate of feature $j$ in frame $i$.

The reprojection residual requires an initial geometric state for the scene and camera trajectory. In practice, an initialization stage is performed to initialize the subsequent spatiotemporal optimization. Specifically, feature tracks are first established across multiple frames using the tracking network. Given these tracks, initial geometric states, including camera poses and feature depths, are estimated through a standard initialization procedure based on triangulation and depth filtering.

The resulting estimates provide the initial state $\mathbf{X}_0$ used to construct reprojection constraints. The detailed initialization process, including depth filtering and track validation, is described in Appendix~\ref{appendix:init}.

The reprojection residual becomes

\begin{equation}
	\mathbf{e}_{ij}(\mathbf{X}, \theta) =
	f_\theta(\mathbf{I}_i, j) -
	\pi(\mathbf{T}_i, \mathbf{X}_j),
	\label{eq:residual_theta}
\end{equation}

where $\mathbf{X}$ is defined as Eq.~(\ref{eq:3}), and
$\pi(\cdot)$ denotes the projection function.
The robust geometric energy is formulated as

\begin{equation}
	E_{\text{reproj}}(\mathbf{X}, \theta) =
	\sum_{(i,j)}
	\rho
	\left(
	\mathbf{e}_{ij}(\mathbf{X}, \theta)^\top
	\mathbf{\Sigma}_{ij}^{-1}
	\mathbf{e}_{ij}(\mathbf{X}, \theta)
	\right),
	\label{eq:joint_energy}
\end{equation}
where $\rho(\cdot)$ is a robust kernel and $\mathbf{\Sigma}_{ij}$ denotes the observation covariance.

The ultimate learning objective is therefore

\begin{equation}
	(\mathbf{X}^*, \theta^*) =
	\arg\min_{\mathbf{X}, \theta}
	E_{\text{reproj}}(\mathbf{X}, \theta).
	\label{eq:joint_opt}
\end{equation}

Directly solving Eq.~(\ref{eq:joint_opt}) by jointly optimizing $(\mathbf{X}, \theta)$ is computationally prohibitive in large-scale robotic systems.A straightforward alternative is to unroll multiple Gauss–Newton or Levenberg–Marquardt iterations and backpropagate through the entire optimization trajectory. However, such unrolling incurs significant memory overhead that scales with the number of iterations, may suffer from numerical instability due to deep computational graphs, and produces gradients that depend on intermediate solver states rather than the fully converged solution, potentially leading to training–inference mismatch.

To overcome these limitations, we follow the philosophy of differentiable nonlinear optimization layers proposed in the Theseus framework~\cite{theseus}. Instead of explicitly differentiating through the optimization trajectory, we treat the geometric solver as an implicit mapping and directly compute gradients of the optimal solution with respect to the network parameters. Formally, the solver is expressed as

\begin{equation}
	\mathbf{X}^*(\theta) = 
	\arg\min_{\mathbf{X}} E_{\text{reproj}}(\mathbf{X}, \theta).
\end{equation}

This formulation corresponds to performing structured second-order optimization on a factor graph, where residuals define factors and states define nodes. During forward propagation, the geometric solver performs iterative local linearization of the nonlinear energy function around the current state estimate. Specifically, at each Gauss--Newton iteration, the local quadratic approximation yields a linear system of the form

\begin{equation}
	\mathcal{H}_k \Delta \mathbf{X}_k = -\mathbf{g}_k,
\end{equation}
where the Hessian matrix and gradient vector are computed at the $k$-th iteration as

\begin{equation}
	\mathcal{H}_k \approx \mathbf{J}_k^\top \mathbf{W}_k \mathbf{J}_k,
	\qquad
	\mathbf{g}_k = \mathbf{J}_k^\top \mathbf{W}_k \mathbf{e}_k.
\end{equation}

Here, $\mathbf{J}_k = \frac{\partial \mathbf{e}_k}{\partial \mathbf{X}}$ denotes the Jacobian matrix of residuals with respect to the state variables at the current linearization point, $\mathbf{e}_k$ represents the stacked reprojection residuals, and $\mathbf{W}_k$ is the measurement confidence weighting matrix derived from observation noise statistics.

From a forward propagation perspective, each iteration computes a locally optimal state increment $\Delta \mathbf{X}_k$, and the solver updates the state estimate until convergence. At convergence, the sequence of incremental updates implicitly accumulates to produce the optimal solution

\begin{equation}
	\mathbf{X}^* = \mathbf{X}_0 + \sum_k \Delta \mathbf{X}_k,
\end{equation}
which corresponds to a stationary point of the nonlinear energy function.

At the converged solution, the first-order optimality condition holds:

\begin{equation}
	\nabla_{\mathbf{X}} E_{\text{reproj}}(\mathbf{X}^*, \theta) = \mathbf{0},
	\label{eq:optimality_condition}
\end{equation}

This indicates that the final solution is independent of the specific optimization trajectory. Therefore, the forward optimization process can be regarded as computing the fixed point of the geometric energy minimization problem.

To enable end-to-end learning, we differentiate Eq.~(\ref{eq:optimality_condition}) with respect to $\theta$ and propagate gradients through the optimal solution rather than through intermediate solver iterations:

\begin{equation}
	\frac{d}{d\theta}
	\nabla_{\mathbf{X}} E_{\text{reproj}}(\mathbf{X}^*, \theta) = \mathbf{0}.
\end{equation}

Applying the chain rule yields

\begin{equation}
	\underbrace{
		\nabla_{\mathbf{X}\mathbf{X}}^2 E_{\text{reproj}}(\mathbf{X}^*, \theta)
	}_{\mathcal{H}}
	\frac{d\mathbf{X}^*}{d\theta}
	+
	\nabla_{\mathbf{X}\theta}^2 E_{\text{reproj}}(\mathbf{X}^*, \theta)
	=
	\mathbf{0}.
	\label{eq:implicit_linear}
\end{equation}

Solving this linear system gives the implicit gradient

\begin{equation}
	\frac{d\mathbf{X}^*}{d\theta}
	=
	-
	\mathcal{H}^{-1}
	\nabla_{\mathbf{X}\theta}^2 E_{\text{reproj}}(\mathbf{X}^*, \theta).
	\label{eq:implicit_solution}
\end{equation}

In practice, the inverse Hessian is never computed explicitly. Instead, gradients are obtained by solving a sparse linear system involving $\mathcal{H}$, which shares the same block-sparse structure as the forward bundle adjustment solver. Furthermore, exploiting the Schur complement structure allows efficient elimination of landmark variables, preserving scalability to large-scale robotic systems. Detailed derivation is in appendix \ref{appendix:mixed_hessian}.

Compared to unrolling iterative solvers, the implicit formulation derives gradients directly from the optimality condition of the geometric objective, thereby avoiding trajectory differentiation while ensuring consistency with the fully converged solution. Moreover, implicit differentiation eliminates the need to store intermediate solver states, significantly reducing memory consumption. This property is particularly important for multi-frame robotic systems, where the number of states increases over time.

\begin{figure*}
	\centering
	{\includegraphics[width=0.9\linewidth]{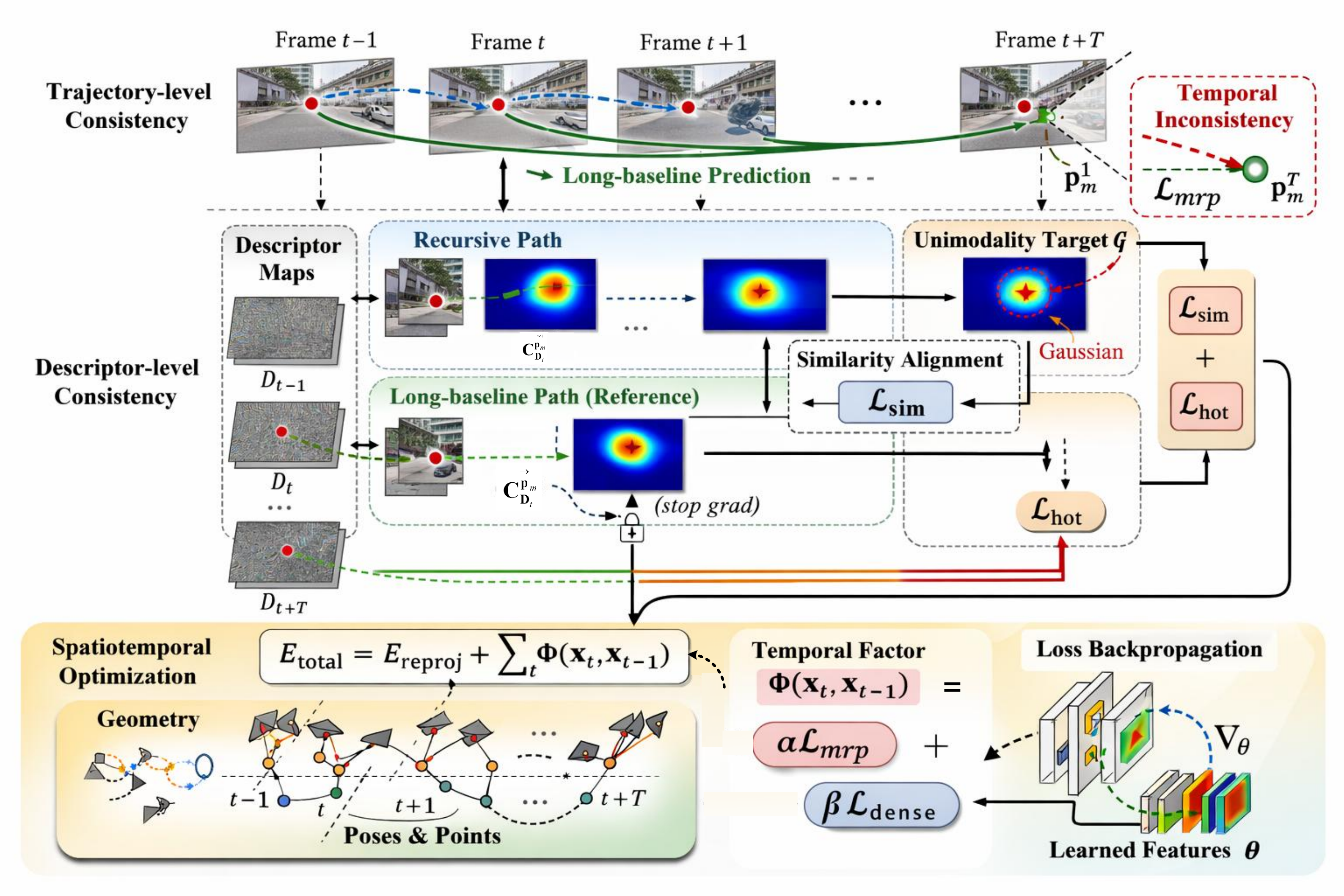}}
	
	\caption{Overall training loss construction pipeline integrating differentiable bundle adjustment with multi-frame temporal consistency at both trajectory and descriptor levels.}
	\label{Fig:4}
\end{figure*}

\subsection{Multi-Frame Temporal Consistency}

While reprojection constraints enforce spatial geometric consistency, visual state estimation in robotics is inherently a spatiotemporal problem, where geometric states evolve continuously across frames. In long video sequences, small estimation errors may accumulate over time, leading to feature trajectory drift and degraded localization stability. To mitigate this issue, we introduce a multi-level temporal consistency formulation that regularizes both geometric states and feature representations across time.

The optimization objective is extended to a spatiotemporal form:

\begin{equation}
	E_{\text{total}}(\mathbf{X},\theta)
	=
	E_{\text{reproj}}(\mathbf{X},\theta)
	+
	\lambda
	\sum_{t\in\mathcal{T}}
	\Phi(\mathbf{x}_t,\mathbf{x}_{t-1}),
\end{equation}
where $E_{\text{reproj}}$ denotes the spatial reprojection energy defined in Eq.~(\ref{eq:joint_energy}). The temporal factor $\Phi(\cdot)$ regularizes state evolution across frames and is constructed from both trajectory-level and descriptor-level consistency constraints.

\paragraph{Multi-frame trajectory consistency.}

We first construct temporal consistency at the trajectory level using feature propagation paths. Consider a feature
point $\mathbf{p}_m^1$ observed in a short video segment $\mathbf{T}=\{1,2,\dots,T\}$.

Recursive tracking propagation is defined as

\begin{equation}
	\overset{\rightsquigarrow}{\mathbf{p}}_m =
	\mathbf{p}_m^1 \rightarrow
	\mathbf{p}_m^2 \rightarrow
	\cdots \rightarrow
	\mathbf{p}_m^T ,
\end{equation}
which corresponds to frame-to-frame sequential estimation.

In contrast, a long-baseline prediction directly estimates the correspondence between the first and last frames

\begin{equation}
	\vv{\mathbf{p}}_m =
	\mathbf{p}_m^1 \rightarrow
	\mathbf{p}_m^T .
\end{equation}

The recursive estimation accumulates local matching uncertainties, whereas the long-baseline estimate provides a
more stable global reference. The discrepancy between these two propagation paths naturally reveals temporal
inconsistency.

The resulting sparse trajectory consistency term is defined as

\begin{equation}
	\mathcal{L}_{mrp}
	=
	\frac{1}{M}
	\sum_{m=1}^{M}
	\operatorname{dist}_{th}
	\left(
	\overset{\rightsquigarrow}{\mathbf{p}}_m,
	\vv{\mathbf{p}}_m
	\right),
\end{equation}
where $M$ denotes the number of valid feature tracks and $\operatorname{dist}_{th}(\cdot)$ removes low-confidence or
outlier correspondences.

\paragraph{Descriptor-level temporal consistency.}

Beyond trajectory-level constraints, we further impose temporal consistency in the descriptor space to stabilize
representation learning. Let $\mathbf{D}_t$ denote the dense descriptor map at frame $t$. Following the two propagation paths, two similarity responses can be computed: $\mathbf{C}^{\overset{\rightsquigarrow}{\mathbf{p}}_m}_{\mathbf{D}_t}$
and $\mathbf{C}^{\vv{\mathbf{p}}_m}_{\mathbf{D}_t}$.

The dense similarity consistency loss is defined as

\begin{equation}
	\mathcal{L}_{sim}
	=
	\left\|
	\left(
	\mathbf{C}^{\overset{\rightsquigarrow}{\mathbf{p}}_m}_{\mathbf{D}_t}
	-
	(\mathbf{C}^{\vv{\mathbf{p}}_m}_{\mathbf{D}_t})_{*}
	\right)_{th}
	\right\|^{2},
\end{equation}
where $(\mathbf{C}^{\vv{\mathbf{p}}_m}_{\mathbf{D}_t})_{*}$ denotes that the cross-frame similarity map is treated as frozen reference and gradients are not propagated through this branch.

While this loss aligns similarity distributions across propagation paths, minimizing it alone may lead to spatially
diffuse responses. To encourage well-localized matches, we further introduce a unimodality constraint.

Specifically, a Gaussian likelihood target centered at the long-baseline prediction is introduced:

\begin{equation}
	\mathbf{G}(x,y)
	=
	\frac{1}{2\pi\sigma^2}
	\exp
	\left(
	-
	\frac{(x-x_m)^2+(y-y_m)^2}{2\sigma^2}
	\right).
\end{equation}

The corresponding unimodality regularization becomes

\begin{equation}
	\mathcal{L}_{hot}
	=
	\left\|
	\left(
	\mathbf{C}^{\overset{\rightsquigarrow}{\mathbf{p}}_m}_{\mathbf{D}_t}
	-
	\mathbf{G}
	\right)_{th}
	\right\|^2.
\end{equation}

Together with the similarity alignment term, this constraint forms a dense descriptor-level supervision:

\begin{equation}
	\mathcal{L}_{dense}
	=
	\mathcal{L}_{sim}
	+
	\mathcal{L}_{hot}.
\end{equation}

Combining sparse trajectory consistency and dense descriptor supervision, the temporal factor can finally be formulated as a multi-level energy:

\begin{equation}
	\Phi(\mathbf{x}_t,\mathbf{x}_{t-1})
	=
	\alpha \mathcal{L}_{mrp}
	+
	\beta \mathcal{L}_{dense},
\end{equation}
where $\alpha$ and $\beta$ balance trajectory stability and descriptor distribution regularization.

\paragraph{Differentiable spatiotemporal optimization.}

Since the temporal factors are incorporated directly into the geometric objective, the implicit differentiation framework remains valid. The Hessian matrix becomes

\begin{equation}
	\mathcal{H}
	=
	\nabla_{\mathbf{X}\mathbf{X}}^2
	E_{\text{total}}(\mathbf{X}^*,\theta),
\end{equation}
which captures both spatial reprojection constraints and multi-frame temporal consistency factors. As illustrated in Fig.~\ref{Fig:4}, the overall training pipeline is constructed within this unified spatiotemporal optimization framework. Consequently, gradients can propagate through the entire optimization graph, enabling end-to-end learning of geometry-aware feature representations.

\section{Implementation details}\label{VI}
\noindent\textbf{Pretraining Recipe:} ViBA is pretrained using real-world geometric relationships as supervision. The model is first pretrained on the MegaDepth dataset, which contains approximately 1 million crowdsourced images covering 196 tourist landmarks, along with camera calibrations and poses obtained via SfM reconstruction, and dense depth maps generated through multi-view stereo (MVS). This pretraining endows the model with initial geometric reasoning capabilities, which are crucial for its generalization to diverse real-world scenarios.

\noindent\textbf{Training Tricks:} Although ViBA features a lightweight architecture with fast inference and high accuracy, we found that certain training details have a significant impact on performance. Since individual videos have limited scene diversity, and the formal training process does not rely on camera poses, depth, or scene annotations, we leverage multi-source videos from the internet as additional training data. Specifically, the model is fine-tuned on the SpatialVID dataset, which contains approximately 2.7 million video clips, totaling around 7,089 hours of dynamic content.

During preprocessing, we observed that some scenes cannot be reliably initialized using standard VIO pipelines (e.g., due to small parallax or highly dynamic objects). Therefore, we retain only video clips with initial geometric projection errors below a specified threshold as training samples, while the remaining clips are labeled as hard samples. These hard samples are selectively included in subsequent training rounds to further enhance the model’s 

\section{EXPERIMENTS}\label{VII}
We evaluate ViBA for the tasks of homography estimation, relative pose estimation, and visual localization. We also analyze the impacts of our design decisions.

\subsection{Homography estimation}
We evaluate our method on the HPatches dataset \cite{hpatches}, which provides planar image sequences with illumination and viewpoint changes along with ground-truth homographies. Since ViBA is designed for local image matching rather than full Structure-from-Motion (SfM) reconstruction, we restrict our evaluation to the subset of HPatches sequences with illumination changes.

Following SuperGlue \cite{superglue}, we report precision and recall under a reprojection error threshold of 3 pixels. For each image pair, homographies are estimated using both a non-robust Direct Linear Transform (DLT \cite{DLT}) solver and a robust RANSAC \cite{RANSAC} estimator. We compute the mean reprojection error of the four image corners and report the area under the cumulative error curve (AUC) at thresholds of 1 px and 5 px. To ensure a fair comparison, the inlier threshold of each method is individually tuned to achieve its best performance.

We compare ViBA against a range of feature extraction and matching pipelines, including nearest-neighbor matching with mutual check (NN + mutual), SuperPoint \cite{SuperPoint} + LightGlue \cite{lightglue}, and recent learned methods such as ALIKED \cite{aliked}, LET-NET \cite{let-net}, ZippyPoint \cite{zippypoint}, and XFeat \cite{xfeat}. For fairness, all sparse methods extract up to 1024 keypoints per image using their official pretrained models. We further evaluate a BA-refined variant for each baseline by feeding initial correspondences into our differentiable bundle adjustment (BA) module.

As shown in Table~\ref{tab:matcher_comparison}, implicit BA consistently improves the precision of correspondences by enforcing geometric consistency and suppressing outliers, while maintaining comparable recall. This leads to more accurate homography estimation, especially for DLT, where the refined correspondences enable a simple non-robust solver to approach the performance of robust estimators. Moreover, the consistent gains in AUC@1 px and AUC@5 px across all baselines further demonstrate the effectiveness of implicit BA in improving geometric accuracy.

\begin{table}[htbp]
	\centering
	\caption{\textbf{Homography estimation on HPatches.} ViBA produces highly reliable correspondences, achieving top precision (P) and high recall (R), enabling accurate downstream tasks such as homography estimation with RANSAC or direct linear transform (DLT).}
	\setlength{\tabcolsep}{4pt} 
	\begin{tabular}{l l c c | c c | c c} 
		\hline
		\multirow{2}{*}{} & \multirow{2}{*}{\centering\textbf{Method}}  & \multirow{2}{*}{\textbf{R}} & \multirow{2}{*}{\textbf{P}} 
		& \multicolumn{2}{c|}{\textbf{AUC - RANSAC}} & \multicolumn{2}{c}{\textbf{AUC - DLT}} \\
		& & & & @1px & @5px & @1px & @5px \\
		\hline
		\multicolumn{8}{c}{\textbf{Features + Matcher}} \\
		\multirow{4}{*}{\rotatebox[origin=c]{90}{\textit{SuperPoint}}} 
		& NN+mutual             & 71.3 & 62.4 & 32.1 & 75.3 & 0.0 & 2.0 \\
		& SuperGlue             & 89.7 & 83.4 & 35.2 & 77.1 & 31.6 & 73.6 \\
		& LightGlue             & 92.6 & 84.9 & 35.7 & 78.5 & 32.3 & 74.0 \\
		& \textbf{ViBA($f_\theta$)} & 90.5 & \textbf{86.8} & \textbf{40.6} & \textbf{84.7} & 35.8 & \textbf{78.5} \\
		\hline
		\multicolumn{8}{c}{\textbf{End-to-End Methods}} \\
		\multirow{4}{*}{\rotatebox[origin=c]{90}{\textit{End-to-End}}} 
		& ALIKED         & 91.2 & 85.7 & 38.5 & 82.7 & 35.6 & 78.3 \\
		& LET-NET        & 89.3 & 78.4 & 33.8 & 79.6 & 29.6 & 72.1 \\
		& ZippyPoint     & 87.6 & 82.2 & 37.6 & 81.8 & 35.4 & 76.6 \\
		& XFeat     & \textbf{93.3} & 84.3 & 37.7 & 80.4 & \textbf{40.6} & 77.8 \\
		\hline
	\end{tabular}
	\label{tab:matcher_comparison}
\end{table}

\begin{table}[b]
	\centering
	\caption{\textbf{Wide-baseline indoor pose estimation.} We report the AUC of pose errors, GPU time (ms), and matching precision (P), all in percent except for time. When applied to the SuperPoint, ViBA outperforms other learning-based matchers. Moreover, in end-to-end evaluations, ViBA achieves the highest efficiency, despite a modest increase in runtime.}
	\setlength{\tabcolsep}{8pt}
	\begin{tabular}{l l c c c c c}
		\hline
		\multirow{2}{*}{} & \multirow{2}{*}{\centering\textbf{Method}} & \multicolumn{3}{c}{\textbf{Pose estimation AUC}} & \multirow{2}{*}{\centering\textbf{P}} & \multirow{2}{*}{\centering\textbf{time}} \\
		\cline{3-5}
		&  & @2° & @5° & @10° &  &  \\
		\hline
		\multicolumn{7}{c}{\textbf{Features + Matcher}} \\
		\multirow{4}{*}{\rotatebox[origin=c]{90}{\textit{SuperPoint}}}  
		& NN + mutual & 23.2 & 43.6 & 67.2 & 60.6 & 4.6 \\
		& SuperGlue & 35.5 & 71.1 & 85.6 & 76.9 & 68.3 \\
		& LightGlue & 36.8 & 75.0 & 86.3 & 77.4 & 44.5 \\
		& \textbf{ViBA($f_\theta$)} & \textbf{42.9} & \textbf{78.8} & 89.1 & \textbf{81.6} & 11.5 \\
		\hline
		\multicolumn{7}{c}{\textbf{End to End Methods}} \\
		\multirow{4}{*}{\rotatebox[origin=c]{90}{\textit{End to End}}}
		& Aliked & 31.2 & 73.4 & 83.6 & 74.5 & 9.7 \\
		& LET-NET & 40.7 & 77.6 & \textbf{90.2} & 81.4 & 12.6 \\
		& ZippyPoint & 36.6 & 72.5 & 84.5 & 78.3 & 196 \\
		& XFeat & 28.7 & 61.8 & 73.7 & 70.2 & 8.6 \\
		\hline
	\end{tabular}
	\label{tab:pose_auc}
\end{table}
\subsection{Indoor pose estimation}
Indoor image matching is highly challenging due to sparse textures, abundant self-similar regions, complex 3D geometries, and significant viewpoint variations. In this work, we demonstrate that ViBA effectively leverages learned local image features to address these challenges.

We evaluate our approach on the 7-Scenes dataset \cite{7scenes}, a small-scale indoor dataset consisting of real RGB-D video sequences with ground-truth camera poses, commonly used for indoor RGB-D odometry and localization research. Owing to the high frame rate, we first select frame pairs suitable for pose estimation based on the ground-truth poses. Subsequently, SuperPoint features are extracted, and ViBA is compared against various baseline matchers. We further evaluate the end-to-end algorithms in terms of pose accuracy. Continuous feature tracking is performed at the original resolution across all sequences, and relative poses are computed for the selected frame pairs. Following prior works [33, 71, 7], we report the area under the curve (AUC) of pose errors at thresholds of 2°, 5°, and 10°, where the pose error is defined as the maximum of rotational and translational errors. Relative poses are estimated via the essential matrix using RANSAC. Additionally, we report matching precision @3px [18] and computational efficiency, with ViBA performing inference on 32×32 image patches.

\begin{table*}[b]
	\centering
	\renewcommand{\arraystretch}{1.2}
	\setlength{\tabcolsep}{1.pt}
	\begin{threeparttable}
		\caption{Absolute Trajectory Error (ATE) and Absolute Rotation Error (ARE). 
			Translation errors are reported in meters, and rotation errors are reported in degrees.
			Red denotes the best, while blue denotes the second best.}
		\label{table:vio}
		
		\begin{tabular}{lc|cc|cc|cc|cc|cc|cc||cc}
			\toprule
			\multirow{2}{*}{} 
			&\multirow{2}{*}{\textbf{Method}} 
			& \multicolumn{2}{c|}{\textbf{MH\_03}}
			& \multicolumn{2}{c|}{\textbf{MH\_05}}
			& \multicolumn{2}{c|}{\textbf{V1\_02}}
			& \multicolumn{2}{c|}{\textbf{V1\_03}}
			& \multicolumn{2}{c|}{\textbf{V2\_02}}
			& \multicolumn{2}{c|}{\textbf{V2\_03}} 
			& \multicolumn{2}{c}{\textbf{Mean}}  \\
			\cmidrule(lr){3-4}
			\cmidrule(lr){5-6}
			\cmidrule(lr){7-8}
			\cmidrule(lr){9-10}
			\cmidrule(lr){11-12}
			\cmidrule(lr){13-14}
			\cmidrule(lr){15-16}
			& & t\_ape (m) & r\_ape (°)
			& t\_ape (m) & r\_ape (°)
			& t\_ape (m) & r\_ape (°)
			& t\_ape (m) & r\_ape (°)
			& t\_ape (m) & r\_ape (°)
			& t\_ape (m) & r\_ape (°)
			& t\_ape (m) & r\_ape (°) \\
			\midrule
			& OpenVINS \cite{openvins}
			& \textcolor{blue}{\textbf{0.1509}} & \textcolor{blue}{\textbf{1.8826}}
			& 0.4936 & 2.2101
			& 0.0583 & \textcolor{red}{\textbf{2.3035}}
			& \textcolor{blue}{\textbf{0.0601}} & 3.0555
			& \textcolor{blue}{\textbf{0.0578}} & \textcolor{blue}{\textbf{2.0987}}
			& 0.2001 & 2.7169
			& 0.1701 & 2.3779 \\
			\hdashline
			\multirow{4}{*}{\rotatebox[origin=c]{90}{\textit{SuperPoint}}}  
			& NN + mutual
			& 0.3125 & 3.2147
			& 0.6821 & 3.8425
			& 0.1184 & 3.9562
			& 0.1567 & 4.2158
			& 0.1432 & 3.4876
			& 0.3985 & 3.9021
			& 0.3019 & 3.7698 \\
			
			& SuperGlue\cite{superglue}
			& 0.1826 & 2.1458
			& 0.4013 & 1.8642
			& 0.0725 & 2.6453
			& 0.0896 & 2.7431
			& 0.0814 & 2.4187
			& 0.2367 & 2.3314
			& 0.1774 & 2.3581  \\
			
			& LightGlue\cite{lightglue}
			& 0.1647 & 2.0279
			& 0.4339 & \textcolor{blue}{\textbf{1.2753}}
			& 0.0709 & 2.8605
			& 0.0920 & \textcolor{blue}{\textbf{2.5322}}
			& 0.0737 & 2.3279
			& 0.2077 & 2.2267
			& 0.1738 & \textcolor{blue}{\textbf{2.2084}} \\
			
			& \textbf{ViBA($f_\theta$)}
			& \textcolor{red}{\textbf{0.1392}} & \textcolor{red}{\textbf{1.8571}}
			& \textcolor{blue}{\textbf{0.3038}} & \textcolor{red}{\textbf{1.1031}}
			& \textcolor{red}{\textbf{0.0537}} & \textcolor{blue}{\textbf{2.3767}}
			& \textcolor{red}{\textbf{0.0597}} & \textcolor{red}{\textbf{1.9230}}
			& \textcolor{red}{\textbf{0.0464}} & 2.3165
			& \textcolor{blue}{\textbf{0.1357}} & \textcolor{red}{\textbf{1.7220}}
			& \textcolor{red}{\textbf{0.1231}} & \textcolor{red}{\textbf{1.8831}} \\
			
			\hdashline
			\multirow{5}{*}{\rotatebox[origin=c]{90}{\textit{End-to-End}}}  
			& ALIKED~\cite{aliked}      
			& 0.1534 & 2.4034
			& \textcolor{red}{\textbf{0.2705}} & 1.3029
			& \textcolor{blue}{\textbf{0.0553}} & 2.5894
			& 0.0749 & 3.0517
			& 0.0670 & \textcolor{red}{\textbf{2.0453}}
			& \textcolor{red}{\textbf{0.1266}} & 2.2436
			& \textcolor{blue}{\textbf{0.1246}} & 2.2727 \\
			
			& LET-NET~\cite{let-net}
			& 0.1958 & 2.2749
			& 0.3481 & 1.2943
			& 0.0875 & 2.4295
			& 0.0847 & 3.4756
			& 0.0657 & 2.2186
			& 0.5872 & \textcolor{blue}{\textbf{2.2181}}
			& 0.2282 & 2.3290 \\
			
			& ZippyPoint~\cite{zippypoint}  
			& 0.2053 & 2.5637
			& 0.4895 & 2.1146
			& 0.0832 & 2.9824
			& 0.1098 & 3.1457
			& 0.0976 & 2.7543
			& 0.2681 & 2.6189
			& 0.2089 & 2.6966  \\
			
			& XFeat~\cite{xfeat}  
			& fail & fail
			& 0.6939 & 3.6493
			& 0.0767 & 2.7285
			& 0.1427 & 2.9422
			& 0.2137 & 4.7513
			& fail & fail
			& 0.2818 & 3.5178 \\
			
			\bottomrule
			\toprule
			\multirow{2}{*}{} 
			& \multirow{2}{*}{\textbf{Method}} 
			& \multicolumn{2}{c|}{\textbf{class\_csc1}}
			& \multicolumn{2}{c|}{\textbf{class\_eng}}
			& \multicolumn{2}{c|}{\textbf{conference\_csc1}}
			& \multicolumn{2}{c|}{\textbf{fantasy\_csc1}}
			& \multicolumn{2}{c|}{\textbf{gattaca\_csc1}}
			& \multicolumn{2}{c|}{\textbf{lab\_module\_csc}}
			& \multicolumn{2}{c}{\textbf{Mean}} \\
			\cmidrule(lr){3-4}
			\cmidrule(lr){5-6}
			\cmidrule(lr){7-8}
			\cmidrule(lr){9-10}
			\cmidrule(lr){11-12}
			\cmidrule(lr){13-14}
			\cmidrule(lr){15-16}
			& & t\_ape (m) & r\_ape (°)
			& t\_ape (m) & r\_ape (°)
			& t\_ape (m) & r\_ape (°)
			& t\_ape (m) & r\_ape (°)
			& t\_ape (m) & r\_ape (°)
			& t\_ape (m) & r\_ape (°)
			& t\_ape (m) & r\_ape (°) \\
			\midrule
			& OpenVINS \cite{openvins}
			& 0.0782 & 17.0680
			& 0.1592 & 46.8445
			& 0.2920 & 43.1039
			& 1.9482 & 142.5431
			& 0.1937 & 17.7539
			& 0.0134 & 17.8477
			& 0.4475 & 47.5269 \\
			\hdashline
			\multirow{4}{*}{\rotatebox[origin=c]{90}{\textit{SuperPoint}}}  
			& NN + mutual
			& 0.0824 & 28.5317
			& 0.1743 & 52.4186
			& 0.3562 & 61.2043
			& 2.7431 & 168.5321
			& 0.2847 & 31.2654
			& 0.2156 & 42.3187
			& 0.6427 & 64.0451  \\
			
			& SuperGlue\cite{superglue}
			& 0.0512 & 19.7423
			& 0.0946 & 14.3187
			& 0.2215 & 45.9321
			& 0.5124 & 63.8475
			& 0.1438 & 12.7543
			& 0.1327 & 21.3489
			& 0.1927 & 29.6573  \\
			
			& LightGlue\cite{lightglue}
			& 0.0468 & 17.2325
			& \textcolor{red}{\textbf{0.0520}} & \textcolor{red}{\textbf{7.3361}}
			& \textcolor{red}{\textbf{0.1898}} & \textcolor{blue}{\textbf{40.4545}}
			& \textcolor{blue}{\textbf{0.3700}} & \textcolor{blue}{\textbf{41.2528}}
			& \textcolor{red}{\textbf{0.0767}} & \textcolor{blue}{\textbf{5.6705}}
			& 0.1470 & 17.1108
			& \textcolor{blue}{\textbf{0.1471}} & \textcolor{blue}{\textbf{21.5095}} \\
			
			& \textbf{ViBA($f_\theta$)}
			& \textcolor{red}{\textbf{0.0204}} & \textcolor{red}{\textbf{4.5848}}
			& 0.0928 & \textcolor{blue}{\textbf{12.8898}}
			& \textcolor{blue}{\textbf{0.1998}} & \textcolor{red}{\textbf{32.7750}}
			& \textcolor{red}{\textbf{0.2288}} & \textcolor{red}{\textbf{35.7748}}
			& 0.1886 & 9.1676
			& \textcolor{red}{\textbf{0.0103}} & \textcolor{blue}{\textbf{3.7803}}
			& \textcolor{red}{\textbf{0.1234}} & \textcolor{red}{\textbf{16.4954}} \\
			
			\hdashline
			\multirow{5}{*}{\rotatebox[origin=c]{90}{\textit{End-to-End}}}  
			& ALIKED~\cite{aliked}      
			& \textcolor{blue}{\textbf{0.0281}} & \textcolor{blue}{\textbf{8.0499}}
			& \textcolor{blue}{\textbf{0.0628}} & 24.5949
			& 0.2749 & 47.5270
			& 1.3561 & 55.7266
			& \textcolor{blue}{\textbf{0.1025}} & \textcolor{red}{\textbf{3.6646}}
			& \textcolor{blue}{\textbf{0.0112}} & \textcolor{red}{\textbf{1.7753}}
			& 0.3059 & 23.5564 \\
			
			& LET-NET~\cite{let-net}
			& 0.2134 & 59.5626
			& fail & fail
			& 0.4265 & 73.8894
			& fail & fail
			& 0.2614 & 21.9927
			& 0.5293 & 36.5864
			& 0.3576 & 48.0078 \\
			
			& ZippyPoint~\cite{zippypoint}  
			& 0.0635 & 22.1843
			& 0.1087 & 18.7325
			& 0.2683 & 52.6478
			& 0.8432 & 81.5632
			& 0.1765 & 16.4372
			& 0.1854 & 26.9087
			& 0.2743 & 36.4123  \\
			
			& XFeat~\cite{xfeat}  
			& 0.2105 & 76.9492
			& fail & fail
			& fail & fail
			& fail & fail
			& 14.9470 & 161.7195
			& fail & fail
			& 7.5788 & 119.3344 \\

			\bottomrule
		\end{tabular}
		\begin{tablenotes}
			\footnotesize
			\item[*] t\_ape(m) represents the absolute translation error, r\_ape(°) represents the absolute rotation error, and Mean represents the average value across all sequences.
		\end{tablenotes}
		
	\end{threeparttable}
\end{table*}

Compared to other learning-based matchers, ViBA achieves significantly higher pose accuracy (see Table \ref{tab:pose_auc}) and integrates seamlessly with SuperPoint features. Its superior accuracy over alternative learned matchers indicates a stronger representation capability. Moreover, ViBA produces more correct matches by focusing on local image regions around feature points, reducing the likelihood of mismatches across the entire image. The combination of the two components in ViBA achieves state-of-the-art performance in indoor pose estimation. These components are complementary: repeatable keypoints that satisfy implicit bundle adjustment constraints enable the estimation of more correct matches, even under highly challenging conditions.

\subsection{VIO Trajectory Estimation}
To further evaluate the effectiveness of ViBA in visual-inertial odometry (VIO), we replace the frontend of OpenVINS with different feature matching methods and conduct experiments on multiple datasets. For a fair comparison, the number of detected keypoints is fixed to 300 for all methods.

Table \ref{table:vio} reports the Absolute Trajectory Error (ATE) and Absolute Rotation Error (ARE) across multiple sequences from the EuRoC\cite{euroc} and UMA \cite{uma} datasets. Overall, simple nearest-neighbor matching (NN + mutual) yields the lowest accuracy, while learning-based matchers such as SuperGlue and LightGlue provide clear improvements. In contrast, ViBA consistently achieves the best or second-best performance across most sequences. On average, ViBA reduces the translation error by approximately 25\%–35\% compared to SuperGlue and by 30\%–40\% compared to LightGlue, while also achieving a relative rotation error reduction of around 15\%–25\%.

Notably, although ALIKED demonstrates competitive performance on the EuRoC dataset, its accuracy degrades when transferred to the UMA dataset. This performance drop can be attributed to limited generalization capability, as end-to-end learned models tend to overfit dataset-specific characteristics such as motion patterns and appearance distributions. In contrast, ViBA maintains stable performance across datasets. This robustness stems from its ability to learn geometric consistency and spatio-temporal consistency directly from arbitrary video streams, enabling better adaptability to diverse environments.

Furthermore, when combined with SuperPoint, ViBA exhibits strong robustness across different motion patterns and scene complexities. Compared with end-to-end methods (e.g., ALIKED, LET-Net, and ZippyPoint), ViBA remains highly competitive and often achieves superior accuracy across multiple sequences. This suggests that decoupling feature extraction and matching, while enforcing stronger geometric constraints, leads to more reliable correspondences and improved pose estimation.

\begin{figure}
	\centering
	{\includegraphics[width=0.9\linewidth]{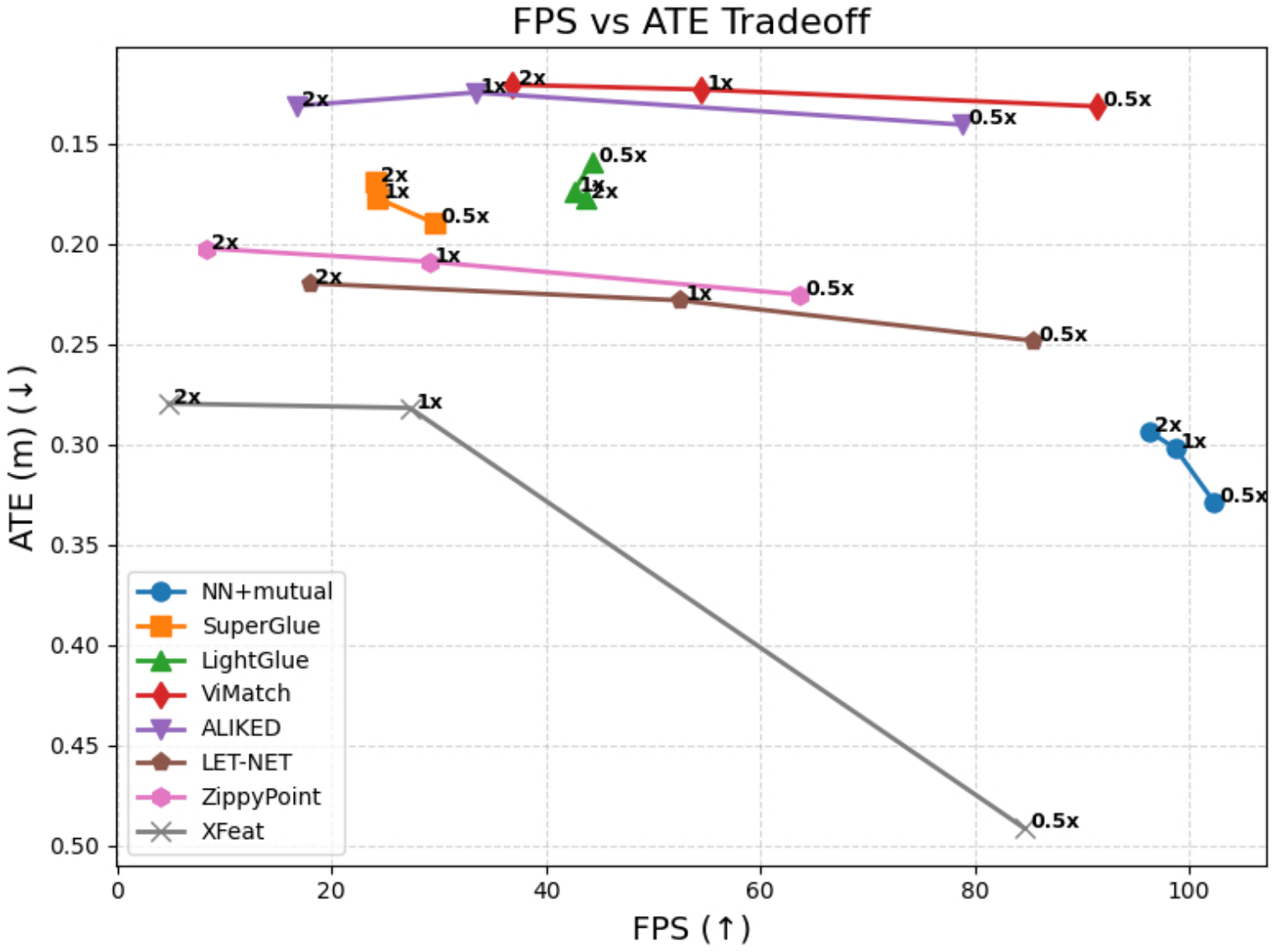}}
	
	\caption{Trade-offs between accuracy and runtime at varying input resolutions.}
	\label{Fig:resolution}
\end{figure}

We further analyze the impact of input resolution on VIO performance, as shown in Table \ref{tab:resolution} and visualized in Fig. \ref{Fig:resolution}. Here, “0.5×” denotes downsampled images at half the original resolution, “1× (origin)” corresponds to the native input resolution, and “2×” indicates images upsampled to twice the original resolution. Since the number of keypoints is fixed, increasing the image resolution does not significantly affect the computational cost of graph-based matchers such as SuperGlue and LightGlue, resulting in relatively stable runtime (FPS) across different resolutions. However, their pose accuracy improves only marginally and remains consistently lower than that of ViBA. In contrast, ViBA benefits more from higher-resolution inputs, achieving an additional 2\%–5\% improvement in ATE while maintaining competitive efficiency. This advantage arises because ViBA explicitly operates on local image patches centered around keypoints, allowing it to better exploit finer image details without increasing matching ambiguity.

Overall, these results demonstrate that ViBA achieves a superior balance between accuracy and efficiency in VIO systems. By focusing on local regions and incorporating implicit bundle adjustment constraints, it produces more reliable correspondences, leading to consistently improved pose estimation performance across diverse datasets and imaging conditions.

\begin{table*}[htbp]
	\centering
	\caption{\textbf{Impact of input resolution on VIO performance.} 
		We report the mean ATE (m) , ARE(°) and runtime FPS on EuRoC sequences using OpenVINS.}
	\setlength{\tabcolsep}{5pt}
	\begin{tabular}{l l l ccc ccc ccc}
		\toprule
		\multirow{2}{*}{} 
		& \multirow{2}{*}{\textbf{Method}} 
		&\multirow{2}{*}{\textbf{Params/M}} 
		& \multicolumn{3}{c}{\textbf{ATE (m)$\downarrow$}} 
		& \multicolumn{3}{c}{\textbf{ARE (°)$\downarrow$}} 
		& \multicolumn{3}{c}{\textbf{FPS$\uparrow$}} \\
		\cmidrule(lr){4-6} \cmidrule(lr){7-9} \cmidrule(lr){10-12}
		
		& & & 0.5x & origin(1x) & 2x & 0.5x & origin(1x) & 2x & 0.5x & origin(1x) & 2x \\
		\midrule
		\multirow{4}{*}{\rotatebox[origin=c]{90}{\textit{SuperPoint}}}  
		& NN + mutual & 1.30 & 0.3285 & 0.3019 & 0.2937 & 3.4215 & 3.7698 & 3.2054 & 102.3 & 98.7 & 96.3 \\
		& SuperGlue   & 13.32 & 0.1897 & 0.1774 & 0.1689 & 2.4821 & 2.3581 & 2.1053 & 29.6 & 24.3 & 24.1 \\
		& LightGlue   & 13.15 & 0.1592 & 0.1738 & 0.1769 & 2.0212 & 2.2084 & \textbf{1.8216} & 44.3 & 42.6 & \textbf{43.7} \\
		& \textbf{ViBA($f_{\theta}$)} & 1.34 & \textbf{0.1315} & \textbf{0.1231} & \textbf{0.1207} & \textbf{1.9543} & \textbf{1.8831} & 1.8320 & \textbf{91.4} & \textbf{54.5} & 36.9 \\
		\hdashline
		\multirow{4}{*}{\rotatebox[origin=c]{90}{\textit{End-to-End}}}  
		& ALIKED     & 0.68 & 0.1406 & 0.1246 & 0.1312 & 2.4801 & 2.2727 & 2.2510 & 78.8 & 33.4 & 16.7 \\
		& LET-NET    & 0.08 & 0.2483 & 0.2282 & 0.2197 & 2.5561 & 2.3290 & 2.1102 & 85.4 & 52.5 & 17.9 \\
		& ZippyPoint & 5.23 & 0.2254 & 0.2089 & 0.2023 & 2.8564 & 2.6966 & 2.5123 & 63.7 & 29.1 & 8.3 \\
		& XFeat      & 1.54 & 0.4912 & 0.2818 & 0.2796 & 5.4831 & 3.5178 & 3.2679 & 84.6 & 27.3 & 4.8 \\
		\bottomrule
	\end{tabular}
	\label{tab:resolution}
\end{table*}

\section{Conclusion}\label{VIII}

In this paper, we present an implicit bundle adjustment (BA) framework that unifies geometric consistency and spatio-temporal consistency within a hybrid optimization paradigm. Unlike conventional approaches that rely on explicit multi-view optimization or purely learned correspondences, the proposed method embeds geometric constraints directly into the learning process, enabling the network to implicitly enforce multi-view consistency while preserving temporal coherence across video sequences.

By jointly modeling spatial reprojection relationships and temporal continuity, the proposed framework effectively reduces mismatches and stabilizes correspondence estimation under challenging conditions such as viewpoint changes, motion blur, and low-texture regions. The implicit formulation further allows the model to leverage dense supervision signals from video streams without requiring explicit pose optimization during inference, resulting in a favorable trade-off between computational efficiency and geometric accuracy.

Overall, this hybrid consistency-driven design provides a principled way to bridge learning-based matching and geometric optimization. It demonstrates that enforcing structured constraints in both spatial and temporal domains is critical for robust correspondence estimation, and offers a scalable solution for real-world visual perception systems. Future work will explore extending this formulation to incorporate global constraints and long-range dependencies for improved consistency at the scene level.


\section{Appendix}

\subsection{Geometric Initialization}
\label{appendix:init}

To bootstrap the spatiotemporal optimization, an initialization stage is performed to estimate initial camera poses and feature depths. This stage ensures that reprojection constraints are well-defined and that depth estimates are numerically stable.

\paragraph{Anchor and candidate frame selection.} 
We fix the first frame of a short window as the anchor frame. Among the remaining frames, the candidate terminal frame is selected based on maximizing co-visibility and parallax while minimizing average reprojection error. Let $F = \{f_1, \dots, f_T\}$ denote the set of frames in the window. For each candidate frame $f_t$, feature correspondences with the anchor frame are established and filtered according to:
\begin{itemize}
	\item Minimum number of correspondences: $N_\text{min}$
	\item Minimum parallax: $\theta_\text{min}$
	\item Maximum allowed reprojection error: $\epsilon_\text{max}$
	\item Minimum positive depth ratio: $r_\text{min}$
\end{itemize}
The frame $f_{t^*}$ that satisfies all constraints and achieves the optimal combination of parallax and reprojection error is selected as the terminal frame.

\paragraph{Relative pose estimation.} 
For the anchor-terminal frame pair $(f_0, f_{t^*})$, normalized image coordinates are computed as
\begin{equation}
	\mathbf{r}_{i,j} = \frac{K^{-1} [\mathbf{p}_{i,j}^\top, 1]^\top}{\| K^{-1} [\mathbf{p}_{i,j}^\top, 1]^\top \|},
\end{equation}
where $\mathbf{p}_{i,j}$ is the pixel coordinate of feature $i$ in frame $j$ and $K$ is the camera intrinsic matrix. The relative rotation $R$ and translation $t$ are recovered from the essential matrix using the five-point algorithm with RANSAC:
\begin{equation}
	\{R, t\} = \text{recoverPose}(E, \mathbf{r}_{i,0}, \mathbf{r}_{i,t^*}).
\end{equation}

\paragraph{Triangulation and depth initialization.} 
Given the relative pose, 3D points are triangulated for each valid correspondence $(\mathbf{p}_{i,0}, \mathbf{p}_{i,t^*})$:
\begin{equation}
	\mathbf{X}_i = \arg \min_{\mathbf{X}} \| \mathbf{p}_{i,0} \times (R_0 \mathbf{X} + t_0) \|^2 + 
	\| \mathbf{p}_{i,t^*} \times (R_{t^*} \mathbf{X} + t_{t^*}) \|^2.
\end{equation}
Points with depths outside the range $[0.1 z_\text{median}, 10 z_\text{median}]$ are discarded, where $z_\text{median}$ is the median depth of valid points in the anchor frame.

\paragraph{Inverse depth update.} 
For each remaining feature, the inverse depth is updated via Bayesian fusion:
\begin{equation}
	\sigma_i^2 = \left(\frac{1}{\sigma_\text{prior}^2} + \frac{1}{\sigma_\text{obs}^2}\right)^{-1}, \quad
	\mu_i = \sigma_i^2 \left(\frac{\mu_\text{prior}}{\sigma_\text{prior}^2} + \frac{\rho_\text{obs}}{\sigma_\text{obs}^2}\right),
\end{equation}
where $\rho_\text{obs} = 1 / z_\text{tri}$ is the observed inverse depth and $\sigma_\text{obs}$ is derived from the pixel-level reprojection error. Features with posterior variance below a threshold are marked as stable for subsequent optimization.

\paragraph{PnP initialization of remaining frames.} 
For each frame $f_j$ in the window (excluding anchor and terminal frames), stable features $(\mathbf{X}_i, \mathbf{p}_{i,j})$ are used to estimate the camera pose via PnP:
\begin{equation}
	P_j = \arg \min_{P} \sum_i \| \mathbf{p}_{i,j} - \pi(P \mathbf{X}_i) \|^2,
\end{equation}
where $\pi(\cdot)$ denotes the projection operation. If PnP fails or produces a large jump, a linear prediction from previous frames is applied:
\begin{equation}
	P_j = P_{j-1} P_{j-2}^{-1} P_{j-1}.
\end{equation}

\paragraph{Sliding window refinement.} 
The triangulation and inverse depth update are repeated for each new frame in the window, maintaining a consistent anchor reference. Stable features are continuously tracked, and inverse depth uncertainty is updated to reflect accumulated observations.  

\begin{figure*}
	\centering
	{\includegraphics[width=0.9\linewidth]{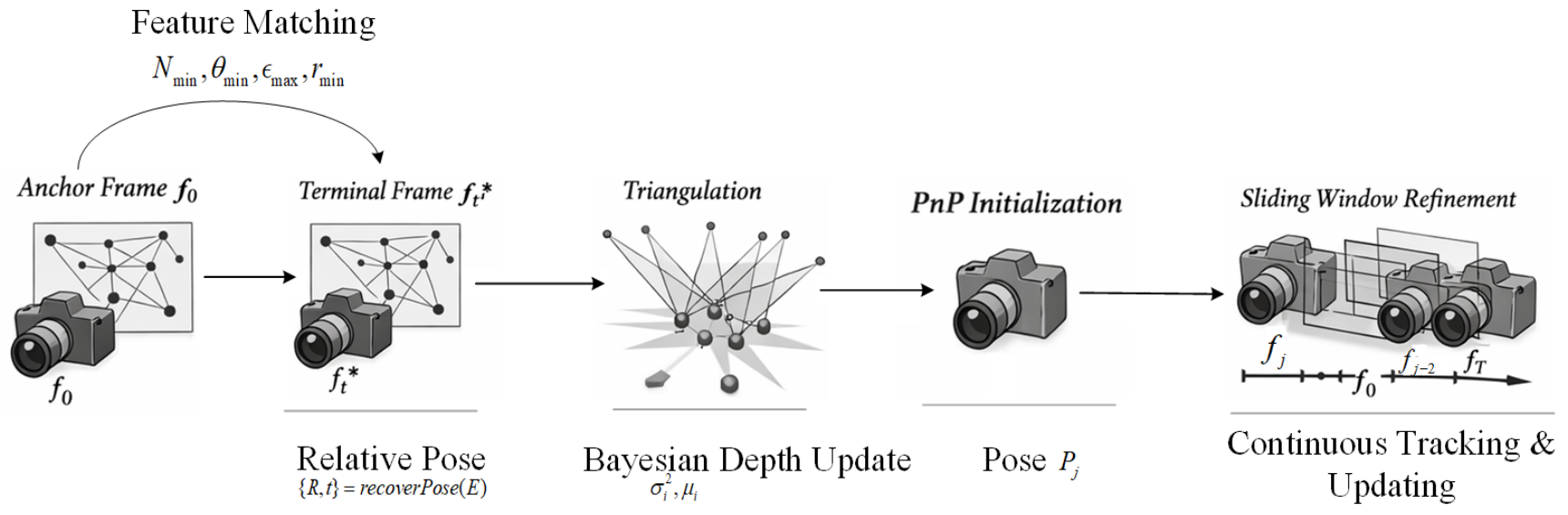}}
	
	\caption{Overview of the geometric initialization process. An anchor frame and a terminal frame are selected to estimate the relative pose and triangulate initial 3D landmarks. Inverse depths are refined through Bayesian updates, after which the poses of remaining frames are initialized via PnP and refined within a sliding window.}
	\label{Fig:init}
\end{figure*}

\paragraph{Resulting initial state.} 
The above procedure produces the initial global state
\begin{equation}
	\mathbf{X} =
	[T_1, T_2, \cdots, T_{N_i},
	p_1, p_2, \cdots, p_{N_j}]^\top ,
\end{equation}
which includes the estimated camera poses and the associated feature depths. As illustrated in Fig.~\ref{Fig:init}, this initialization provides the starting point for the reprojection-based optimization described in Sec.~\ref{BA}.

\subsection{Gradient Derivation of the Mixed Hessian}
\label{appendix:mixed_hessian}

The mixed Hessian term is defined as

\begin{equation}
	\nabla_{\mathbf{X}\theta}^2 E_{\text{reproj}}
	=
	\frac{\partial}{\partial \theta}
	\left(
	\frac{\partial E_{\text{reproj}}}{\partial \mathbf{x}}
	\right),
\end{equation}
which corresponds to first computing the gradient of the energy with respect to the geometric state $\mathbf{x}$ and then differentiating it with respect to the network parameters $\theta$.

The bundle adjustment objective can be written as

\begin{equation}
	E_{\text{reproj}}(\mathbf{X},\theta)
	=
	\sum_{i,j}
	\left\|
	\mathbf{e}_{ij}(\mathbf{X},\theta)
	\right\|^2,
\end{equation}
where the reprojection residual is defined as

\begin{equation}
	\mathbf{e}_{ij}(\mathbf{X},\theta)
	=
	\pi(\mathbf{T}_i,\mathbf{P}_j)
	-
	\hat{\mathbf{p}}_{ij}(\theta).
\end{equation}

Here

\begin{itemize}
	\item $\mathbf{x}=(\mathbf{T},\mathbf{P})$ denotes the geometric state variables, including camera poses and landmark positions,
	\item $\pi(\mathbf{T}_i,\mathbf{P}_j)$ represents the projection function,
	\item $\hat{\mathbf{p}}_{ij}(\theta)$ is the feature observation predicted by the tracking network.
\end{itemize}

The gradient of the energy function with respect to the geometric state can be written as

\begin{equation}
	\frac{\partial E_{\text{reproj}}}{\partial \mathbf{X}}
	=
	2
	\sum_{i,j}
	\mathbf{J}_{x,ij}^{\top}
	\mathbf{e}_{ij},
\end{equation}
where

\begin{equation}
	\mathbf{J}_{x,ij}
	=
	\frac{\partial \mathbf{e}_{ij}}{\partial \mathbf{X}}
\end{equation}
is the Jacobian of the residual with respect to the geometric variables.

We now differentiate the above expression with respect to the network parameters $\theta$:

\begin{equation}
	\frac{\partial}{\partial \theta}
	\left(
	\mathbf{J}_{x}^{\top}\mathbf{e}
	\right)
	=
	\left(
	\frac{\partial \mathbf{J}_{x}}{\partial \theta}
	\right)^{\top}\mathbf{e}
	+
	\mathbf{J}_{x}^{\top}
	\frac{\partial \mathbf{e}}{\partial \theta}.
\end{equation}

In bundle adjustment, the residual is defined as

\begin{equation}
	\mathbf{e}
	=
	\pi(\mathbf{x})
	-
	\hat{\mathbf{p}}(\theta).
\end{equation}

Since the projection function $\pi(\mathbf{x})$ depends only on the geometric variables and is independent of the network parameters $\theta$, the Jacobian

\begin{equation}
	\mathbf{J}_x
	=
	\frac{\partial \mathbf{e}}{\partial \mathbf{x}}
	=
	\frac{\partial \pi}{\partial \mathbf{x}}
\end{equation}

does not depend on $\theta$. Therefore,

\begin{equation}
	\frac{\partial \mathbf{J}_x}{\partial \theta}
	=
	0.
\end{equation}

Consequently, the mixed Hessian simplifies to

\begin{equation}
	\nabla_{\mathbf{X}\theta}^2 E_{\text{reproj}}
	=
	\mathbf{J}_x^{\top}
	\frac{\partial \mathbf{e}}{\partial \theta}.
\end{equation}

Using the residual definition

\begin{equation}
	\mathbf{e}
	=
	\pi(\mathbf{x})
	-
	\hat{\mathbf{p}}(\theta),
\end{equation}

its derivative with respect to the network parameters becomes

\begin{equation}
	\frac{\partial \mathbf{e}}{\partial \theta}
	=
	-
	\frac{\partial \hat{\mathbf{p}}}{\partial \theta}.
\end{equation}

Substituting into the mixed Hessian yields

\begin{equation}
	\nabla_{\mathbf{X}\theta}^2 E_{\text{reproj}}
	=
	-
	\mathbf{J}_x^{\top}
	\frac{\partial \hat{\mathbf{p}}}{\partial \theta}.
\end{equation}

From the implicit differentiation result in Eq.~(\ref{eq:implicit_solution}), the gradient of the loss with respect to the network parameters is

\begin{equation}
	\frac{dL}{d\theta}
	=
	-
	\frac{\partial L}{\partial \mathbf{X}^*}
	\mathcal{H}^{-1}
	\nabla_{\mathbf{X}\theta}^2 E_{\text{reproj}}.
\end{equation}

Substituting the derived expression of the mixed Hessian gives

\begin{equation}
	\frac{dL}{d\theta}
	=
	\frac{\partial L}{\partial \mathbf{X}^*}
	\mathcal{H}^{-1}
	\mathbf{J}_x^{\top}
	\frac{\partial \hat{\mathbf{p}}}{\partial \theta}.
\end{equation}

This result shows that gradients from the geometric optimization layer are propagated to the tracking network through the reprojection residuals and the bundle adjustment Jacobian. The term $\partial \hat{\mathbf{p}} / \partial \theta$ corresponds to the Jacobian of the tracking network and can be efficiently computed using automatic differentiation.

\bibliographystyle{unsrt}
\bibliography{ref}

\end{document}